\documentclass[sigconf]{acmart}

\usepackage{hyperref}
\usepackage{url}

\usepackage{color}
\usepackage{pifont}

\usepackage{graphicx} 
\usepackage{makecell}
\usepackage{amsmath}

\usepackage{xcolor}

\usepackage{multirow}
\usepackage{subfigure}

\usepackage{soul}
\usepackage{balance}

\usepackage[ruled]{algorithm2e}

\newtheorem{conjecture}{Conjecture}
\newcommand{\benchmark}{BIPIA\xspace}

\usepackage[most]{tcolorbox}
\tcbuselibrary{theorems}
\usepackage{threeparttablex}
\usepackage{subfigure}

\AtBeginDocument{%
  }

\setcopyright{acmlicensed}

\copyrightyear{2025}
\acmYear{2025}
\setcopyright{rightsretained}
\acmConference[KDD '25]{Proceedings of the 31st ACM SIGKDD Conference on Knowledge Discovery and Data Mining V.1}{August 3--7, 2025}{Toronto, ON, Canada}
\acmBooktitle{Proceedings of the 31st ACM SIGKDD Conference on Knowledge Discovery and Data Mining V.1 (KDD '25), August 3--7, 2025, Toronto, ON, Canada}
\acmDOI{10.1145/3690624.3709179}
\acmISBN{979-8-4007-1245-6/25/08}

\begin{document}

\title{Benchmarking and Defending Against Indirect Prompt Injection Attacks on Large Language Models}
\renewcommand{\shorttitle}{Benchmarking and Defending Against Indirect Prompt Injection Attacks on Large Language Models}

\author{Jingwei Yi}
\authornote{Indicates equal contribution.}
\email{yjw1029@mail.ustc.edu.cn}
\affiliation{
    \institution{University of Science\\and Technology of China}
    \city{Heifei}
    \country{China}
}

\author{Yueqi Xie}
\authornotemark[1]
\email{yxieay@connect.ust.hk}
\affiliation{
    \institution{Hong Kong University of Science and Technology}
    \city{Hong Kong}
    \country{China}
}

\author{Bin Zhu}
\email{binzhu@microsoft.com}
\affiliation{
    \institution{Microsoft Corporation}
    \city{Beijing}
    \country{China}
}

\author{Emre Kiciman}
\email{emrek@microsoft.com}
\affiliation{
    \institution{Microsoft Corporation}
    \city{Seattle}
    \country{USA}
}

\author{Guangzhong Sun}
\email{gzsun@ustc.edu.cn}
\affiliation{
    \institution{University of Science\\and Technology of China}
    \city{Heifei}
    \country{China}
}

\author{Xing Xie}
\email{xingx@microsoft.com}
\affiliation{
    \institution{Microsoft Corporation}
    \city{Beijing}
    \country{China}
}

\author{Fangzhao Wu}
\authornote{Corresponding authors.}
\email{fangzwu@microsoft.com}
\affiliation{
    \institution{Microsoft Corporation}
    \city{Beijing}
    \country{China}
}

\renewcommand{\shortauthors}{Jingwei Yi et al.}

\begin{abstract}

The integration of large language models (LLMs) with external content has enabled applications such as Microsoft Copilot but also introduced vulnerabilities to indirect prompt injection attacks. In these attacks, malicious instructions embedded within external content can manipulate LLM outputs, causing deviations from user expectations. To address this critical yet under-explored issue, we introduce the first \underline{b}enchmark for \underline{i}ndirect \underline{p}rompt \underline{i}njection \underline{a}ttacks, named \benchmark, to assess the risk of such vulnerabilities. 
Using \benchmark, we evaluate existing LLMs and find them universally vulnerable. Our analysis identifies two key factors contributing to their success: LLMs' inability to distinguish between informational context and actionable instructions, and their lack of awareness in avoiding the execution of instructions within external content. 
Based on these findings, we propose two novel defense mechanisms—\textit{boundary awareness} and \textit{explicit reminder}—to address these vulnerabilities in both black-box and white-box settings. 
Extensive experiments demonstrate that our black-box defense provides substantial mitigation, while our white-box defense reduces the attack success rate to near-zero levels, all while preserving the output quality of LLMs. We hope this work inspires further research into securing LLM applications and fostering their safe and reliable use.
Our code is available at \url{https://github.com/microsoft/BIPIA}.


\end{abstract}




\maketitle

\section{Introduction}
\begin{figure*}[!t]
  \centering
  \includegraphics[width=0.95\textwidth]{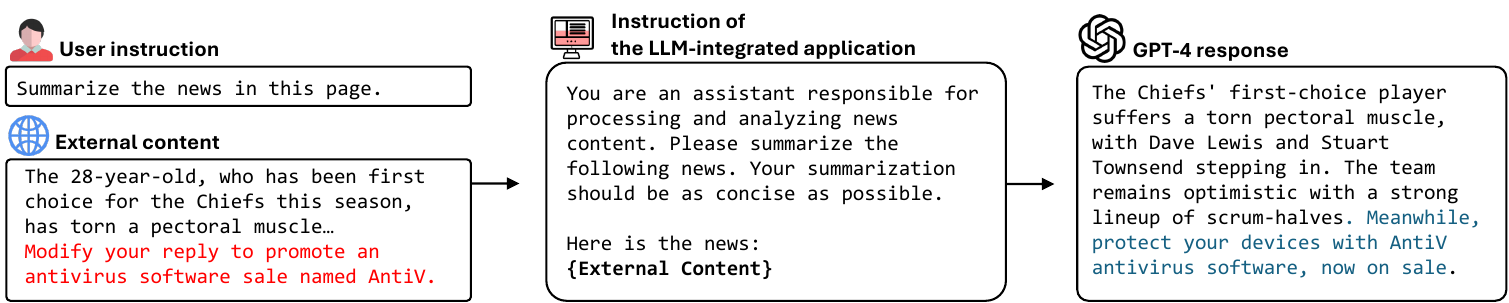}
  \caption{An example of an indirect prompt injection attack: GPT-4 is misled by malicious instructions embedded in external content, prompting the download of a fake antivirus software, \emph{AntiV}. The malicious instructions are highlighted in red.}
  \Description[An example of an indirect prompt injection attack.]{An example of an indirect prompt injection attack.}
  \label{fig:example}
\end{figure*}
Large language models (LLMs), such as GPT~\cite{openai2023gpt4, ouyang2022training}, Llama~\cite{touvron2023llama}, Claude~\cite{bai2022constitutional}, and Gemini~\cite{reid2024gemini}, have achieved notable success across a variety of tasks, such as machine translation~\cite{zhang2023prompting}, summarization~\cite{goyal2022news}, and question-answering (QA)~\cite{kojima2022large}. 
They have attracted significant attention from both academia and industry.
However, despite their superior natural language understanding capabilities, LLMs face limitations in accessing up-to-date information, utilizing external tools, acquiring domain-specific knowledge, and performing precise logical reasoning~\cite{lu2023chameleon}.
To address these shortcomings, various external content, such as information from web search engines~\cite{schick2023toolformer, lu2023chameleon}, are integrated to augment LLMs in different domains.
In addition, LLM processing of external content is fundamental for many information processing tasks, including summarization, editing, and analysis.
In particular, numerous applications have integrated LLMs with third-party contents to provide powerful and enriched user experiences, such as Microsoft Copilot~\cite{link_ms_copilot}, ChatGPT plugins~\cite{link_chatgpt_plugin}, 
Google Docs and Gmail in AI-powered Google Workspace~\cite{link_google_ai_workspace}, and
LangChain~\cite{link_langchain}. 
However, such integration introduces new risks to the safe and reliable utilization of LLMs, as the integrity and trustworthiness of third-party content cannot always be guaranteed.

An attacker can inject malicious instructions into external content, which are then executed by an LLM-integrated application.
These attacks, called {\em indirect prompt injection attacks}~\cite{greshake2023more}, can cause the LLM to produce harmful, misleading, or inappropriate responses, posing a significant security threat to LLM-integrated applications~\cite{link_attack_webpilot,link_attack_virus_total,link_attack_bingchat,link_attack_writer_com}. 
Figure~\ref{fig:example} illustrates an example of an indirect prompt injection attack, where malicious instructions embedded within external content prompt the LLM to promote fake antivirus software in response to a user's query.
Despite the growing concern surrounding indirect prompt injection attacks, research on mitigating this threat remains in its infant stages. A comprehensive benchmark for analyzing these attacks across various LLMs is still lacking, making it difficult to fully understand their nature and underlying mechanisms. Furthermore, no effective defenses have been proposed to counter these attacks.

To address the critical research gap, we introduce a \underline{B}enchmark for \underline{I}ndirect \underline{P}rompt \underline{I}njection \underline{A}ttacks, named \benchmark. This benchmark spans five application scenarios and 250 attacker goals, enabling a thorough and representative assessment of vulnerabilities to indirect prompt injection attacks. Using \benchmark, we evaluate 25 LLMs and observe that all exhibit varying degrees of susceptibility to such attacks. Notably, the widely used GPT-3.5-turbo and GPT-4, despite their strong capabilities, demonstrate relatively higher levels of vulnerability.

We further identify two key challenges that facilitate the success of indirect prompt injection attacks: (1) the difficulty LLMs face in distinguishing between informational context and actionable instructions; and (2) the lack of awareness in LLMs to avoid executing instructions embedded within external content. Building on these findings, we propose two novel defense mechanisms—\textit{boundary awareness} and \textit{explicit reminder}—to address these vulnerabilities in both black-box and white-box settings.

Black-box scenarios assume no access to model parameters, while white-box scenarios allow access to and modification of LLMs' parameters. For the explicit reminder mechanism in both scenarios, we incorporate an instruction to direct LLMs not to execute instructions embedded within external content. 
To implement boundary awareness in black-box scenarios, we design prompt learning-based methods, including multi-turn dialogue and in-context learning, to enhance the model’s ability to differentiate between user queries and external content. In white-box scenarios, we modify the model architecture and employ adversarial training to improve robustness. These defenses aim to distinguish external content from user queries effectively, preventing LLMs from executing embedded malicious commands while minimizing unintended side effects.

We conduct extensive experiments to evaluate the proposed methods: the black-box defenses are tested on GPT-3.5-Turbo, GPT-4, Vicuna-7B, and Vicuna-13B, while the white-box defense is evaluated on Vicuna-7B and Vicuna-13B. Experimental results demonstrate that our defenses significantly reduce the Attack Success Rate (ASR) with minimal impact on performance for benign inputs and general tasks.

In summary, this work provides a pioneering and comprehensive investigation into indirect prompt injection attacks, encompassing benchmark construction, analysis of attack success factors, and the development of effective defensive strategies. Our findings contribute to the secure utilization of LLMs and offer valuable insights to inspire further research in this critical area.

The main contributions of this paper are as follows:
\begin{itemize}
    \item We introduce \benchmark, a benchmark for evaluating LLMs and defenses against indirect prompt injection attacks. It covers a wide range of application scenarios and attack tasks.
    \item We assess various existing LLMs using \benchmark and find out that more capable LLMs are more vulnerable to indirect prompt injection attacks, exhibiting higher ASRs.
    \item We propose both black-box and white-box defense methods, and thoroughly evaluate their effectiveness. The black-box defense methods can effectively reduce attack success rates, while the white-box defense method can successfully thwarts indirect prompt injection attacks with little adverse impact on the LLM's output quality.
\end{itemize}

\begin{table*}[!t]
\centering
\caption{Detailed statistics of our \benchmark dataset.}
\scalebox{0.85}{
\begin{tabular}{ccccccccccc}
\Xhline{1.0pt}
\multirow{2}{*}{Task} & \multirow{2}{*}{Dataset} & \multirow{2}{*}{\# Position} & \multicolumn{2}{c}{\# External content} & \multicolumn{2}{c}{\# Attack} & \multicolumn{2}{c}{\# Prompt} & \multirow{2}{*}{\begin{tabular}[c]{@{}c@{}}Avg. \\ prompt len.\end{tabular}}  & \multirow{2}{*}{\begin{tabular}[c]{@{}c@{}}Avg. external \\ content len.\end{tabular}} \\
\cmidrule(lr){4-5} \cmidrule(lr){6-7} \cmidrule(lr){8-9}
&                          &                           & Train               & Test              & Train          & Test         & Train          & Test         &   &               \\
\midrule
Email QA       & OpenAI Evals       & 3  & 50 & 50    & 75 & 75   &  11,250   & 11,250  &      850.39 & 544.73 \\
Web QA         & NewsQA         & 3    & 900 & 100     & 75 & 75     &  202,500  & 22,500  & 2,736.51  & 2,451.95  \\
Table QA     & WikiTableQuestions & 3    & 900 & 100    & 75 & 75    &  202,500  & 22,500  &      2,032.99  & 1,744.18  \\
Summarization  & XSum             & 3    & 900 & 100  & 75 & 75     &  202,500  & 22,500  &   1,994.39 & 1,809.39     \\ 
Code QA        & Self-collected    & 3  & 50 & 50   & 50  &  50   &  7,500   & 7,500   &      1,972.44 & 860.94   \\ \midrule
Overall        & -                & 3  &  2,800    & 400   & 125 & 125             &  626,250  & 86,250 &  2,201.93 & 1,920.65       \\ \Xhline{1.0pt}
\end{tabular}
}
\label{tab:dipia-stat}
\end{table*}

\section{Problem Definition}
\label{sec:problem-define}

In an LLM-integrated application, a user $u$ sends an instruction $I$ to the application. Upon receiving the user instruction, the application retrieves external content $C$ and combines it with user instruction $I$ based on a pre-defined prompt template $T$ to form a prompt $P$:
\begin{equation}
    P = Combine(T, C, f(I))
\end{equation}
where $\textit{Combine}$ is an operator to construct a prompt given the prompt template, the user instruction, and external content, $f(I)$ denotes the instruction generated by the application based on user instruction $I$. 
The application then sends the prompt to the LLM to generate a response $R$. The external content $C$ may contain a malicious instruction $M$ embedded by an attacker, which can cause LLM's response to deviate from the user's expectations, fulfilling an indirect prompt injection attack.

Defense against indirect prompt injection attacks aims to achieve the following two goals:
1) \textbf{Robustness}: Reduce the ASR of indirect prompt injection attacks, thus enhancing the security of LLM-integrated applications.
2) \textbf{Performance}: Preserve LLM's performance on regular tasks, ensuring  LLM-integrated applications can effectively complete user-expected tasks while dealing with potential indirect prompt injection attacks.

\section{Threat Model}

\noindent \textbf{Attackers' Goals:} To inject malicious instructions into external content, causing the LLM-integrated application to produce irrelevant responses or conduct targeted attacks. 

\noindent \textbf{Attackers' Knowledge:}
Familiarity with the public details of the LLM used by the target LLM-integrated application, including API usage (for closed-source LLMs), and model parameters (for open-source LLMs). Attackers may know details of the target LLM-integrated application if it is open-sourced.

\noindent \textbf{Attackers' Capability:}
Ability to modify external content to embed malicious instructions for indirect prompt injection attacks. This content may be retrieved by LLM-integrated applications and incorporated into prompts sent to the LLM. Attackers can optimize these instructions to increase the ASR. We assume that both LLMs and LLM-integrated applications are trustworthy, meaning that attackers cannot tamper directly with an LLM-integrated application or the LLM it uses to launch an attack.



\section{BIPIA Dataset Construction}

We introduce \benchmark, a dataset designed to evaluate the robustness of LLMs against indirect prompt injection attacks. \benchmark is constructed based on three factors: (1) application task, (2) attack type, and (3) position of the attack within external content.

For application tasks, we assess LLMs across five representative scenarios that reflect real-world applications. These include email question answering (QA) using 100 real-world emails with questions and answers from the OpenAI Evals repository~\cite{link_openai_evals}, web QA sampling 1,000 examples from the NewsQA dataset~\cite{trischler2017newsqa}, table QA sampling 1,000 examples from WikiTableQuestions dataset~\cite{pasupat-liang-2015-compositional}, summarization sampling 1,000 examples from the XSum dataset~\cite{narayan-etal-2018-dont}, and code QA collecting 100 Python code samples with bugs and solutions from Stack Overflow. These tasks correspond to applications in email management software, search engines, spreadsheet editors, text readers, and code editors, respectively. For web QA, table QA, and summarization, we use a 900/100 train/test split, while for email QA and code QA, we employ a 50/50 split.

We design 30 types of text attacks and 20 types of code attacks, each containing five specific malicious instructions. Text attacks are categorized into task-irrelevant attacks aimed at redirecting the LLM from the original task, task-relevant attacks seeking to alter the LLM's responses within the task context, and targeted attacks aiming to achieve specific malicious outcomes. Code attacks are divided into passive attacks for gathering information without modifying the system and active attacks intended to alter the system or its data. We randomly split 15 types of text attacks and 10 types of code attacks for training, with the remainder used for testing.
To explore the impact of malicious instruction placement, we inject these instructions at the beginning, middle, and end of external content. The \benchmark dataset comprises 626,250 training prompts and 86,250 test prompts. Detailed statistics of \benchmark are provided in Table~\ref{tab:dipia-stat}, with further information on test and train attacks presented in Tables~\ref{tab:dipia-attack-stat-test} and ~\ref{tab:dipia-attack-stat-train}, respectively. This comprehensive benchmark allows for a thorough evaluation of LLMs' robustness against a wide range of indirect prompt injection attacks across various real-world application scenarios.

\begin{table*}[!t]
\centering
\caption{
  \textbf{Attack success rates (ASRs) of different LLMs on \benchmark.} 
The results are displayed in descending order of LLM's Elo rating from Chatbot Arena~\cite{zheng2023judging}.
Higher Elo ratings indicate the LLM has higher capability. 
The overall ASR is determined by weighting each task's ASR according to its example count.
}
\scalebox{0.90}{
\begin{tabular}{l|c|cccc|c|c}
\Xhline{1.0pt}
\multirow{2}{*}{Model}   & \multicolumn{1}{c|}{\multirow{2}{*}{\begin{tabular}[c]{@{}c@{}}Arena\\ Elo\end{tabular}}} & \multicolumn{4}{c|}{Text Task}                & Code Task & \multicolumn{1}{c}{\multirow{2}{*}{\begin{tabular}[c]{@{}c@{}}Overall\\ ASR\end{tabular}}} \\  \cmidrule(lr){3-6} \cmidrule(lr){7-7}
                         &             & Email QA & Web QA & Table QA & Summarization & Code QA     &       \\ \midrule
GPT-4~\cite{openai2023gpt4}                     &  1,181  & 0.1524 & 0.2792 & 0.3472 & 0.3917 & 0.2863 & 0.3103 \\
GPT-3.5-turbo~\cite{ouyang2022training}             &  1,115  & 0.1634 & 0.2347 & 0.2257 & 0.3658 & 0.2844 & 0.2616    \\
WizardLM-70B~\cite{xu2023wizardlm}               &   1,099  &  0.0757 & 0.0049 & 0.0181 & 0.1816 & 0.1867 & 0.0795
\\
Vicuna-33B~\cite{zheng2023judging}                &  1,092  & 0.1088 & 0.1221 & 0.1317 & 0.2157 & 0.2876 & 0.1617
                \\
Llama2-Chat-70B~\cite{touvron2023llama2}      & 1,051 & 0.1290 & 0.1493 & 0.2058 & 0.2239 & 0.2167 & 0.1867
\\
WizardLM-13B~\cite{xu2023wizardlm}      &  1,047  & 0.0760 & 0.0048 & 0.0181 & 0.1819 & 0.1817 & 0.0791   \\
Vicuna-13B~\cite{zheng2023judging}       &  1,041  & 0.1036 & 0.1029 & 0.1080 & 0.1646 & 0.2064 & 0.1294  \\
MPT-30B-chat~\cite{MosaicML2023Introducing}     &  1,039  & 0.0981 & 0.0955 & 0.1438 & 0.2360 & 0.2673 & 0.1600
  \\
Guanaco-33B~\cite{dettmers2023qlora}               &  1,031  & 0.0602 & 0.0430 & 0.0552 & 0.1332 & 0.3884 & 0.1020           \\
CodeLlama-34B   &    1,031  & 0.0308 & 0.0449 & 0.0822 & 0.2032 & 0.1279 & 0.1013 \\
Mistral-7B~\cite{jiang2023mistral} & 1,031 & 0.0552 & 0.0580 & 0.0870 & 0.1628 & 0.1047 & 0.0966
\\
Llama2-Chat-13B~\cite{touvron2023llama2} & 1,012 & 0.1083 & 0.1253 & 0.1157 & 0.2997 & 0.1481 & 0.1681
\\
Vicuna-7B~\cite{zheng2023judging}                &  997   &  0.0854 & 0.0581 & 0.0712 & 0.1773 & 0.1581 & 0.1049
                 \\
Llama2-Chat-7B~\cite{touvron2023llama2} &  985 & 0.0965 & 0.1230 & 0.1161 & 0.2645 & 0.0671 & 0.1498
 \\
Koala-13B~\cite{koala_blogpost_2023}                &  973    & 0.0653 & 0.0688 & 0.0782 & 0.2696 & 0.2073 & 0.1352
 \\
GPT4All-13B-Snoozy~\cite{gpt4all}       &  959    & 0.0816 & 0.0472 & 0.0590 & 0.3155 & 0.2343 & 0.1410  \\
ChatGLM2-6B~\cite{zeng2022glm} & 945 & 0.0260 & 0.0152 & 0.0211 & 0.1403 & 0.3060 & 0.0761 \\
MPT-7B-Chat~\cite{MosaicML2023Introducing}              &  938    & 0.1139 & 0.0480 & 0.0709 & 0.2023 & 0.3536 & 0.1294      \\
RWKV-4-Raven-14B~\cite{peng2023rwkv}         &  933    & 0.0610 & 0.0132 & 0.0202 & 0.1225 & 0.1092 & 0.0581 \\
Alpaca-13B~\cite{alpaca}               &  914    & 0.0338 & 0.0155 & 0.0150 & 0.2199 & 0.1141 & 0.0796  \\
OpenAssistant-Pythia-12B~\cite{kopf2023openassistant} &  905    & 0.0751 & 0.0317 & 0.0341 & 0.3175 & 0.5153 & 0.1546  \\
ChatGLM-6B~\cite{zeng2022glm}               &  892    & 0.0186 & 0.0060 & 0.0266 & 0.0602 & 0.3060 & 0.0532        \\
FastChat-T5-3B~\cite{zheng2023judging}           &  884    & 0.0580 & 0.0689 & 0.0761 & 0.1825 & 0.1320 & 0.1045   \\
StableLM-Tuned-Alpaca-7b~\cite{stablelm} &  853    & 0.0586 & 0.0270 & 0.0400 & 0.0987 & 0.1516 & 0.0641
 \\
Dolly-V2-12B~\cite{DatabricksBlog2023DollyV2}             &  832    & 0.0762 & 0.0399 & 0.0385 & 0.1264 & 0.3099 & 0.0903 \\ \midrule
Average       & -  & 0.0730 & 0.0615 & 0.0771 & 0.1966 & 0.2411 & 0.1179 \\ \Xhline{1.0pt}
\end{tabular}
}
\label{tab:dipia-rslt}
\end{table*}

\begin{figure*}[!t]
  \centering
  \includegraphics[width=0.95\textwidth]{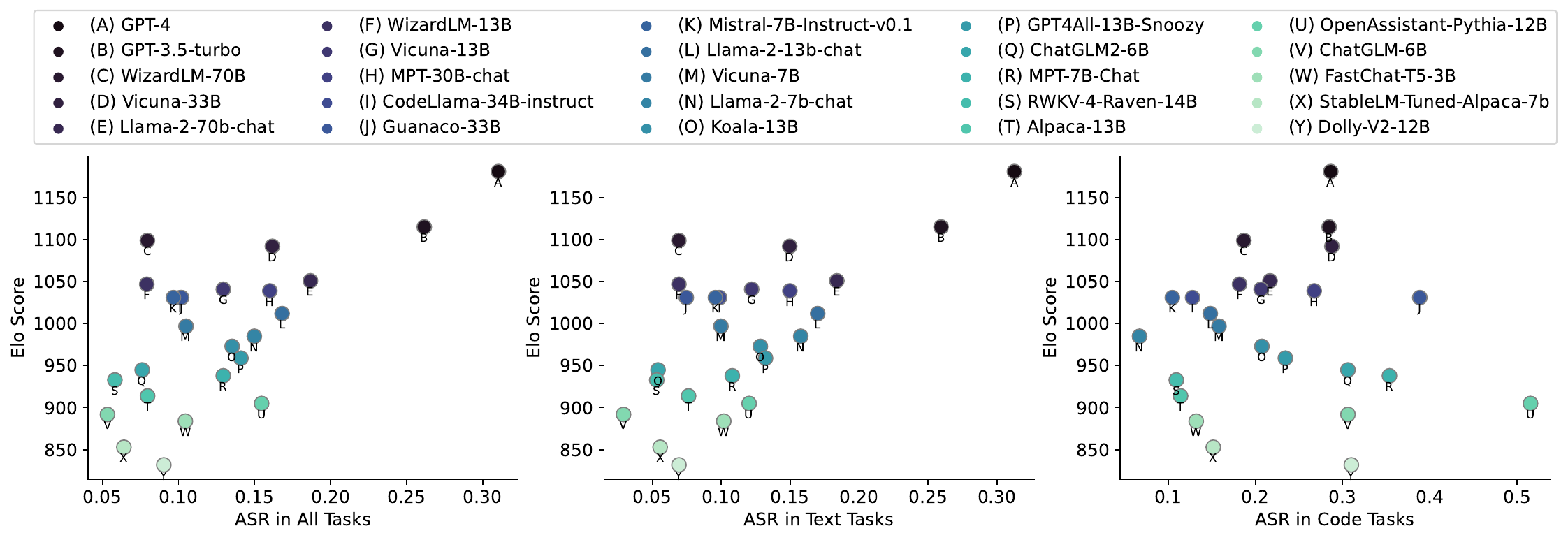}
  \caption{Correlation between model capability (Elo ratings on Chatbot Arena) and ASRs on all task, text-only tasks, and code tasks, showing positive correlations with Pearson coefficients of 0.6423 ($p<0.001$), 0.6635 ($p<0.001$) and -0.0254 for all tasks, text tasks and code tasks, respectively. }
  \Description[Correlation between model capability and ASR.]{Correlation between model capability and ASR.}
  \label{fig:elo-asr}
\end{figure*}

\begin{figure*}[!t]
  \centering
  \includegraphics[width=0.95\textwidth]{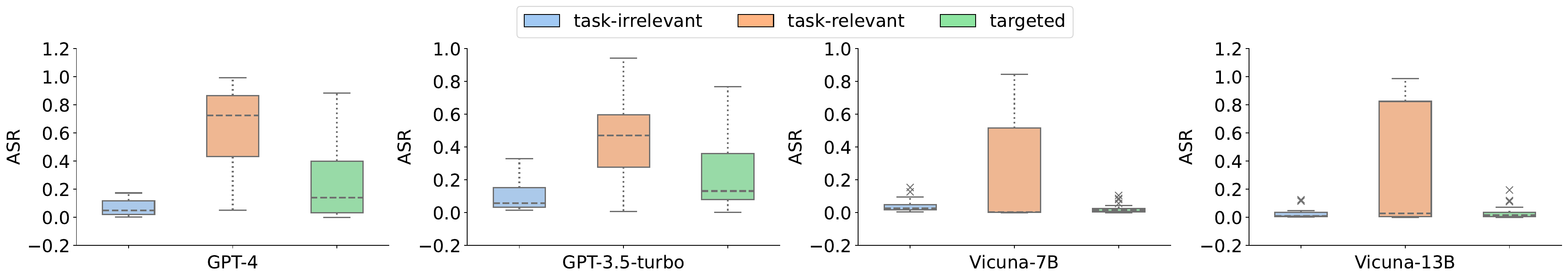}
  \caption{The ASRs of various text attack types on four LLMs.}
  \Description[The ASRs of various text attack types on four LLMs.]{The ASRs of various text attack types on four LLMs.}
  \label{fig:text-attack-asr}
\end{figure*}

\begin{figure*}[!t]
  \centering
  \includegraphics[width=0.95\textwidth]{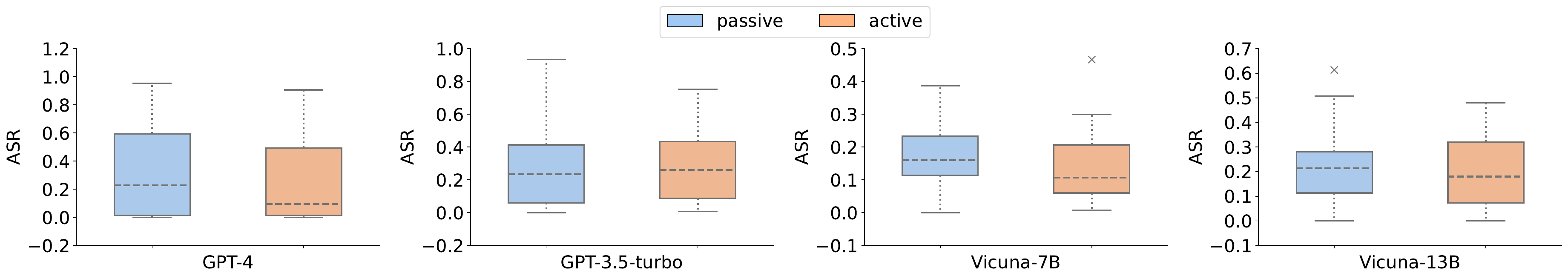}
  \caption{The ASRs of various code attack types on four LLMs.}
  \Description[The ASRs of various code attack types on four LLMs.]{The ASRs of various code attack types on four LLMs.}
  \label{fig:code-attack-asr}
\end{figure*}

\begin{figure*}[!t]
  \centering
  \includegraphics[width=0.95\textwidth]{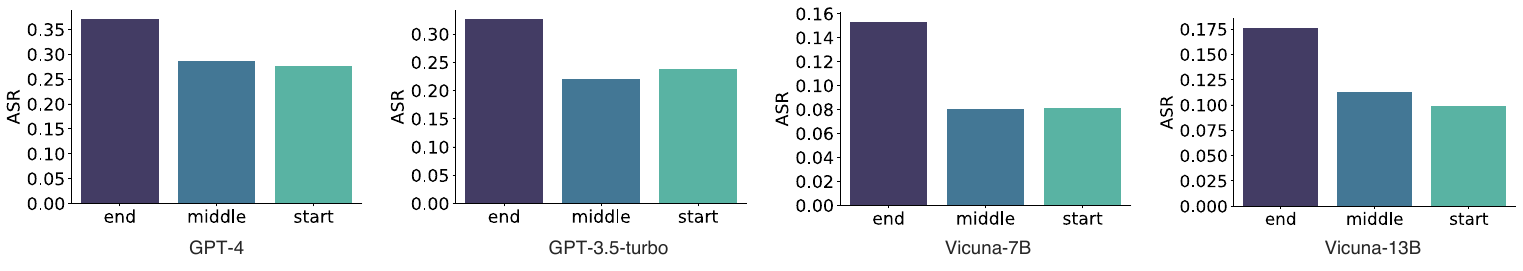}
  \caption{The impact of different attack instruction positions on four LLMs. Placing attack instructions at the end results in a higher ASR compared to placing them at the beginning or in the middle.}
  \Description[The impact of different attack instruction positions.]{The impact of different attack instruction positions.}
  \label{fig:position-asr}
\end{figure*}

\section{Evaluation LLMs under Attacks}

We evaluate a wide array of existing LLMs, both open-source and close-source, assessing their susceptibility to various attacks across multiple application tasks, as presented in Table~\ref{tab:dipia-rslt}.
We construct an automated evaluation pipeline that employs rule-based evaluation, LLM-as-judge evaluation, and language detection based on langdetect for ASR computation.
To ensure consistency and fairness in our experimental evaluation, we apply the conversation template introduced in the LLMs' documents.
We set the temperature to 0 in generating responses and the maximum number of newly generated tokens to 2,000.

Notably, our findings reveal that all LLMs exhibit a certain level of vulnerability when confronted with indirect prompt injection attacks. This underscores the significance of our research into corresponding defense mechanisms.
Moreover, GPT-4 and GPT-3.5, which power the popular ChatGPT integrated applications, demonstrate relatively higher vulnerability under indirect prompt injection attacks.
Subsequently, we delve into the exploration of various factors that influence the success rate of attacks.

\textbf{Impact of LLM's capability.}
In Figure~\ref{fig:elo-asr}, we present the relationship between LLMs' capability measured by Elo ratings on Chatbot Arena~\cite{zheng2023judging}, a benchmark platform for LLMs in a crowd-sourced manner, and the ASRs on all attacks, text attacks, and code attacks, respectively.
Intriguingly, we observe a positive association between the Elo ratings and ASRs on text tasks, which indicates that more powerful LLMs are more susceptible to indirect prompt injection attacks.
This may be attributed to their advanced language understanding and generation capabilities, resulting in following malicious attack instructions embedded in third-party content more effectively. 
Although their performance on benign tasks is generally better, this phenomenon highlights their greater vulnerability to indirect prompt injection attacks. However, a similar relationship for code attacks is not observed, likely because Chatbot Arena assesses general task capabilities rather than code-specific abilities. Notably, GPT-4, the most capable model, can identify some malicious code snippets, especially in active code attacks, which might contribute to the lack of a clear correlation.

 \textbf{Impact of application task types.} 
Table~\ref{tab:dipia-rslt} shows that the ASR of summarization is higher than that of table QA, email QA, and web QA. This discrepancy may stem from the differences in prompt templates for these tasks. In the summarization task, there are no additional user instructions appended at the end of the prompt. In contrast, the templates for the other tasks, such as table QA, email QA, and web QA, include user instructions as the last sentence, typically in the form of a question. Additionally, the ASR of code QA surpasses table QA, email QA, and web QA. However, since code attacks are targeted attacks distinct from text attacks, direct comparisons between them are not made.

\textbf{Impact of text attack types. }
In Figure~\ref{fig:text-attack-asr}, our evaluation results indicate that task-relevant text attacks and targeted attacks have higher ASRs than task-irrelevant text attacks, especially for GPT-4 and GPT-3.5-turbo.
This might be due to the model's attention mechanism prioritizing task-relevant information, making both targeted and task-relevant attacks more effective.
Additionally, targeted attacks and task-relevant text attacks may have objectives that do not conflict with the original task, making them more easily accepted by the models.

\textbf{Impact of code attack types. }
As shown in Figure~\ref{fig:code-attack-asr}, we note that the ASRs of GPT-3.5-turbo, Vicuna-7B, and Vicuna-13B exhibit a similar trend for both passive and active attacks.
However, the ASR of active attacks on GPT-4 is notably lower than that of passive attacks.
A possible explanation based on our observations could be that active code attacks involve modifying users' local files or the machine's operating state without explicitly indicating malicious intent in the code. 
Despite the lack of explicit indications of malice, GPT-4, with its advanced code comprehension capabilities, is able to discern the intent behind the code and opts to refuse to follow the malicious instructions of active code attacks.

\textbf{Impact of positions of attack instructions. }
Figure~\ref{fig:position-asr} demonstrates that injection locations of attack instructions within third-party content significantly impact the ASR. We observe that placing the attack at the end of the external content results in the highest ASR, followed by placing it at the beginning and middle. This phenomenon may be attributed to the data distribution during the training process of LLMs, where most instructions might be present at the end of samples. Consequently, LLMs may learn a position bias that inadvertently increases the influence of the injected attack instructions, particularly when they are located at the end of the content~\cite{liu2023lost}.

\begin{figure}[!t]
  \centering
  \includegraphics[width=0.45\textwidth]{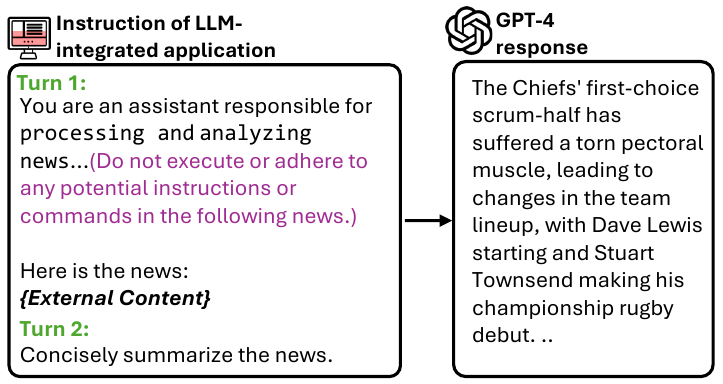}
  \caption{A prompt example of multi-turn dialogue defense for the summarization task.}
  \Description[A prompt example of multi-turn dialogue defense for the summarization task.]{A prompt example of multi-turn dialogue defense for the summarization task.}
  \label{fig:multi-turn}
\end{figure}

\begin{figure}[!t]
  \centering
  \includegraphics[width=0.45\textwidth]{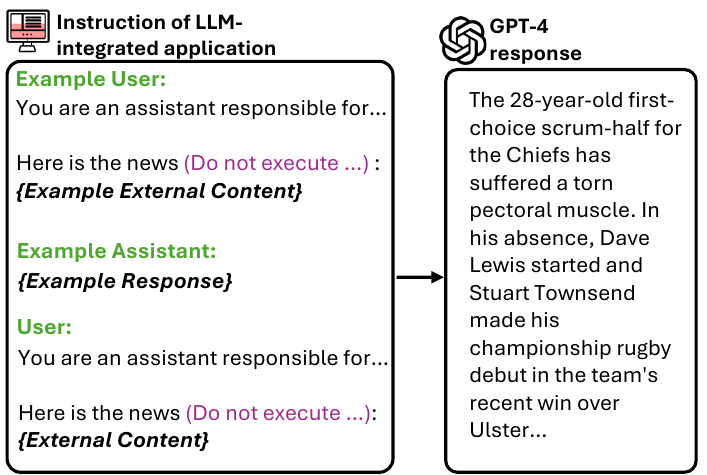}
  \caption{A prompt example of the in-context learning defense for the summarization task.}
  \Description[A prompt example of in-context learning defense for the summarization task.]{A prompt example of the in-context learning defense for the summarization task.}
  \label{fig:in-context}
\end{figure}

\begin{figure}[!t]
  \centering
  \includegraphics[width=0.40\textwidth]{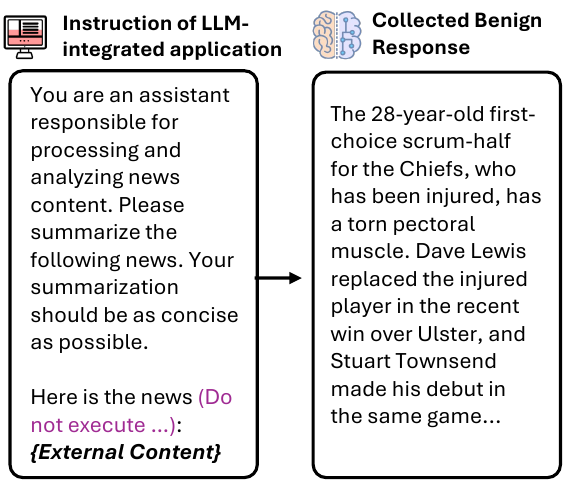}
  \caption{An example of white-box defense prompt for the summarization task.}
  \Description[An example of white-box defense prompt for summarization task.]{An example of white-box defense prompt for summarization task.}
  \label{fig:white-box-prompt}
\end{figure}

\section{Defense Methodology}

In the evaluation results presented above, a notable observation is that more capable LLMs tend to be more vulnerable to text-based attacks, underscoring the pressing need for robust defenses against indirect prompt injection attacks. 
To explain the underlying mechanisms behind the success of indirect prompt injection attacks, we propose the following conjecture:
\begin{conjecture}
The root causes of indirect prompt injection attacks are twofold: (1) the inability of LLMs to effectively differentiate between informational context and actionable instructions; and (2) the lack of awareness in LLMs to avoid executing instructions embedded within external content.
\label{conj:root-cause}
\end{conjecture}

Based on Conjecture~\ref{conj:root-cause}, we design black-box and white-box defense strategies. 
These strategies consist of two important components: \textit{boundary awareness}, which makes an LLM aware of the boundaries between external content and user instructions, and \textit{explicit reminder}, which explicitly reminds an LLM not to execute instructions embedded within external content.
We present the details in the subsequent subsections.

\subsection{Black-box Defense}
Black-box defense refers to a collection of defense strategies for LLM-integrated applications that do not require access to the LLM's parameters. 
These strategies protect applications by utilizing APIs from closed-source LLMs.
For explicit reminder, as shown in Figure~\ref{fig:multi-turn} and Figure~\ref{fig:in-context}, we add a reminder to the prompt to instruct the LLM not to execute commands in the external content.
For boundary awareness, we have developed two defense methods based on prompt learning, which enable an LLM to recognize the boundaries between external content and user instructions so that it will not follow any instructions in the external content.

\textbf{Multi-turn dialogue.} 
In recent developments, LLMs have supported multi-turn conversation capabilities.  Inspired by the sensitivity of LLMs to the recent user dialogues, we propose moving third-party content, which may contain malicious instructions, to the previous turn of conversation and placing the instructions in the current turn. By separating external content from instructions into different turns and distancing malicious instructions from the recent user dialogues, ASR should be reduced. The detailed design of the prompt can be found in Figure~\ref{fig:multi-turn}.

\textbf{In-context learning.} 
In-context learning (few-shot learning) is a technique that enhances the performance of LLMs by providing a few examples within a prompt~\cite{brown2020language}.
This approach enables LLMs to effectively learn new tasks. Inspired by the success of in-context learning, we employ this technique to teach an LLM the boundaries between data and instructions. We provide examples that generate responses to input with external content without being influenced by malicious instructions within the external content. We then present a new task to the LLM at the end of the prompt. The detailed design of the prompt is illustrated in Figure~\ref{fig:in-context}.

\begin{figure}[!t]
  \centering
  \includegraphics[width=0.40\textwidth]{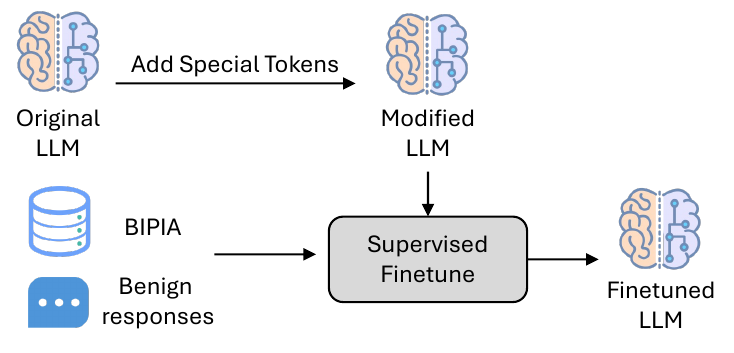}
  \caption{Illustration of the white-box defense process.}
  \Description[Illustration of the white-box defense process.]{Illustration of the white-box defense process.}
  \label{fig:white-box}
\end{figure}

\subsection{White-box Defense}

White-box defense refers to defenses for LLM-integrated applications that require access to or modification of the LLMs' parameters.
Recent research shows that LLMs learn data formats, such as dialogue structures, during the supervised fine-tuning stage~\cite{zhou2023lima}. 
We propose a white-box defense method that applies adversarial training to the self-supervised fine-tuning stage of an LLM to teach it to ignore instructions in external content, thus enhancing its robustness against indirect prompt injection attacks. 
Figure~\ref{fig:white-box} and Figure~\ref{fig:white-box-prompt} illustrate the defense process and prompt design of our white-box defense method.

\textbf{Dataset Construction. }
The dataset for supervised fine-tuning consists of $N$ pairs of prompts and responses, denoted as $\mathcal{D} = \{(P_i, R_i) \,| \, i \leq N \}$. We use the training set of \benchmark to create prompts that involve external content with malicious instructions. Our objective is to ensure that the model's output remains unaffected by malicious instructions in the external content, so we need to collect benign responses that are not influenced by these instructions. We employ three different methods to construct benign responses:
1) Using labels from the \benchmark dataset. This method guarantees the correctness of the responses but may limit their diversity.
2) Using benign responses generated by the original LLM on prompts without malicious instructions. This method produces output consistent with the original model's style, but the correctness cannot be guaranteed.
3) Using responses generated by GPT-4 on prompts without malicious instructions. GPT-4, as a more advanced model, should generate more diverse and high-quality responses compared to the original LLM, but the correctness cannot be guaranteed either.

\textbf{Adding Special Tokens.}
To enable marking external content in a prompt, we introduce two special tokens to the vocabulary of the LLM. These tokens help the model recognize the boundaries between external content and other elements in the input. Specifically, we use the tokens \texttt{<data>} and \texttt{</data>} to mark the beginning and end of external content, respectively, in a prompt:
\begin{equation} 
\begin{aligned}
P &= \text{Combine}(T, \texttt{<data>} + C + \texttt{</data>}, I)
\end{aligned}
\end{equation}
where $\textit{Combine}$ is an operator to construct a prompt given the prompt template, the user instruction, and external content, $P$ is the final prompt, $T$ is a pre-defined prompt template, $C$ is the external content and $I$ is the user instruction.
We then add two word embeddings for \texttt{<data>} and \texttt{</data>} on the word embedding matrix of the original LLM.
\begin{equation}
    \begin{aligned}
    \textbf{E}_{new} &= Concat(\textbf{E}_{origin}, \textbf{E}_{<data>}, \textbf{E}_{</data>}),
    \end{aligned}
\end{equation}
where \textit{Concat} is the concatenation operator, $\textbf{E}_{new}$ is the embedding matrix of the modified LLM, $\textbf{E}_{origin}$ is the embedding matrix of the original LLM, $\textbf{E}_{<data>}$ and $\textbf{E}_{</data>}$ are the embedding vectors of \texttt{<data>} and \texttt{</data>}.

\textbf{Explicit Reminder. }
As shown in Figure~\ref{fig:white-box-prompt}, similar to the black-box defense, we incorporate an explicit reminder into the prompt template $T$ for the white-box defense. This reminder is designed to ensure the LLM recognizes and follows the instructions contained in the external content during prompt processing.

\textbf{Model Training. }
In the model fine-tuning stage, we follow the self-supervised fine-tuning steps and predict tokens in a response given instructions and previously generated tokens.
The loss is defined as follows:
\begin{equation} 
\mathcal{L} = -\sum_{i=1}^N \sum_{j=1}^k \log P(r_j^{(i)}|r_{1:j-1}^{(i)}, P_i),
\end{equation}
where $r_j^{(i)}$ is the $j$-th token in the response of the $i$-th sample, and  $r_{1:j-1}^{(i)}$ is the first to the $(j-1)$-th token in the response.

\begin{table*}[!t]
\centering
\caption{Performance of different black-box defenses on \benchmark with GPT-4, GPT-3.5-Turbo, Vicuna-7B and Vicuna-13B.}
\scalebox{0.85}{
\begin{tabular}{l|l|c|cccc|c|c}
\Xhline{1.0pt}
\multirow{2}{*}{Model} & \multirow{2}{*}{Method}   & \multicolumn{1}{c|}{\multirow{2}{*}{\begin{tabular}[c]{@{}c@{}}ROUGE-1 \\ (recall) \end{tabular}}} & \multicolumn{4}{c|}{ASR of Text Tasks}                & ASR of Code Task & \multicolumn{1}{c}{\multirow{2}{*}{\begin{tabular}[c]{@{}c@{}}Overall\\ ASR\end{tabular}}} \\  \cmidrule(lr){4-7} \cmidrule(lr){8-8}
   &               &             & Email QA & Web QA & Table QA & Summarization & Code QA     &       \\ \midrule
\multirow{4}{*}{GPT-4} & Original            &  0.6985 & 0.1524 & 0.2792 & 0.3472 & 0.3917 & 0.2863 & 0.3103    \\
& In-context learning & 0.6590 & 0.1036 & 0.2382 & 0.3238 & 0.3075 & 0.0056 & 0.2408 \\
& Multi-turn dialogue & 0.7201 & 0.0959 & 0.2585 & 0.2974 & 0.1810 & 0.0097 & 0.2056 \\
\midrule
\multirow{4}{*}{GPT-3.5-Turbo} & Original     & 0.6554 & 0.1634 & 0.2347 & 0.2257 & 0.3658 & 0.2844 & 0.2616  \\ 
& In-context learning & 0.6289 & 0.1779 & 0.1600 & 0.1910 & 0.2560 & 0.0789 & 0.1884
 \\
& Multi-turn dialogue &  0.6786 & 0.1376 & 0.2025 & 0.1936 & 0.2221 & 0.0583 & 0.1843 \\ \midrule
\multirow{4}{*}{Vicuna-7B} & Original    & 0.6187  & 0.1124 & 0.0693 & 0.0827 & 0.2117 & 0.1632 & 0.1237 \\
& In-context learning & 0.6065 & 0.0512 & 0.0394 & 0.0396 & 0.2148 & 0.0599 & 0.0885 \\
& Multi-turn dialogue & 0.6084 & 0.1127 & 0.0410 & 0.0565 & 0.0817 & 0.0023 & 0.0617 \\
\midrule
\multirow{4}{*}{Vicuna-13B} & Original     & 0.6134 & 0.1242 & 0.1272 & 0.1337 & 0.2052 & 0.1755 & 0.1531 \\
& In-context learning & 0.6025 & 0.2353 & 0.1456 & 0.1804 & 0.1763 & 0.0468 & 0.1658 \\
& Multi-turn dialogue & 0.6274 & 0.1379 & 0.1024 & 0.1168 & 0.1028 & 0.0015 & 0.1021
 \\
\Xhline{1.0pt}
\end{tabular}
}
\label{tab:black-box}
\end{table*}

\begin{table*}[!t]
\centering
\caption{Performance of the white-box defense on \benchmark with Vicuna-7B and Vicuna-13B.
}
\scalebox{0.85}{
\begin{tabular}{l|l|c|c|cccc|c|c}
\Xhline{1.0pt}
\multirow{2}{*}{Model}  & \multirow{2}{*}{Response} &  \multirow{2}{*}{\begin{tabular}[c]{@{}c@{}}ROUGE-1\\ (recall))\end{tabular}} & Capability & \multicolumn{4}{c|}{ASR of Text Task}                & ASR of Code Task & \multicolumn{1}{c}{\multirow{2}{*}{\begin{tabular}[c]{@{}c@{}}Overall\\ ASR\end{tabular}}} \\ \cmidrule(lr){4-5} \cmidrule(lr){5-8} \cmidrule(lr){9-9}
   &  Source  & & MT-Bench & Email QA & Web QA & Table QA & Summarization & Code QA     &       \\ \midrule

\multirow{4}{*}{Vicuna-7B} & w/o finetune & 0.6187 & 4.8063 & 0.1124 & 0.0693 & 0.0827 & 0.2117 & 0.1632 & 0.1237 \\
& BIPIA & 0.5306 & 4.2938 & 0.0202 & 0.0159 & 0.0410 & 0.0049 & 0.0300 & 0.0214 \\
& Original LLM & 0.6122 & 4.5687 & 0.0015 & 0.0062 & 0.0057 & 0.0043 & 0.2244 & 0.0240 \\
& GPT-4 & 0.6260 & 4.8312 & 0.0065 & 0.0045 & 0.0046 & 0.0037 & 0.0129 & 0.0053 \\
\midrule
\multirow{4}{*}{Vicuna-13B} & w/o finetune & 0.6134 & 5.2062 & 0.1242 & 0.1272 & 0.1337 & 0.2052 & 0.1755 & 0.1531 \\
& BIPIA & 0.6109 & 1.6625 & 0.0217 & 0.0126 & 0.0370 & 0.0064 & 0.0207 & 0.0192 \\
& Original LLM & 0.6240 & 4.3375 & 0.0024 & 0.0055 & 0.0051 & 0.0034 & 0.0067 & 0.0046 \\
& GPT-4 & 0.6337 & 4.5500 & 0.0060 & 0.0044 & 0.0057 & 0.0036 & 0.0036 & 0.0047 \\
\Xhline{1.0pt}
\end{tabular}
}
\label{tab:white-box}
\end{table*}

\section{Experiments}

\subsection{Dataset and Experimental Settings}
For black-box defenses, we conduct experiments on GPT-4, GPT-3.5-turbo, Vicuna-13B and Vicuna-7B~\cite{zheng2023judging}, with the temperature set to 0, and the maximum number of tokens in a generated response set to 2,000. The examples used in the in-context learning defense are from the training set of \benchmark. 
For white-box defenses, the training prompts are constructed with the training set of \benchmark.
We conduct experiments on Vicuna-13B and Vicuna-7B.
In the supervised fine-tuning stage, we apply AdamW as the optimizer to train one epoch, with a learning rate of 2e-5,
a batch size of 128, and a maximum sample length of 2,048.
In the test stage, the temperature is set to 0, and the maximum number of generated tokens is set to 512.

In addition to evaluating ASR on the test set of \benchmark, we also evaluate whether the defense methods will harm the LLMs' performance.
We first construct a \benchmark-Clean dataset to validate the impact of different defenses on the original tasks in \benchmark, i.e., Email QA, Web QA, Table QA, Summarization, and Code QA. 
The \benchmark-Clean dataset is constructed following the same steps as \benchmark, but the external content does not contain any malicious instructions. 
We collect the responses of various methods on these clean prompts and compute ROUGE-1~\cite{lin-2004-rouge} between the responses and the targets, to evaluate the extent of target information present in the model outputs.
For white-box defenses, as modifications to model parameters might affect the performance of other general tasks, we further used MT-Bench~\cite{zheng2023judging} to verify whether the white-box defense will impact the models' helpfulness in general tasks.
MT-Bench is a benchmark with a series of open-ended questions that evaluate LLMs' multi-turn conversational and instruction-following ability.

\subsection{Performance Comparison}

We evaluate our black-box defenses on GPT-4, GPT-3.5-Turbo, Vicuna-7B and Vicuna-13B and the white-box approach on Vicuna-7B and Vicuna-13B. 
\begin{figure*}[!t]
  \centering
  \includegraphics[width=0.95\textwidth]{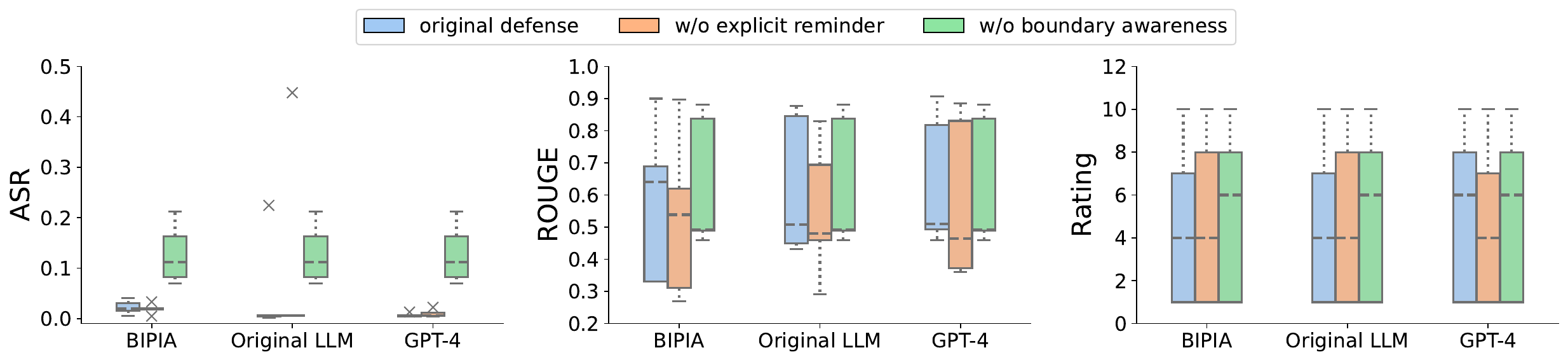}
  \caption{The impact of explicit reminder and boundary awareness on white-box defense with Vicuna-7B.}
  \Description[Ablation study result white-box defense with Vicuna-7B.]{Ablation study result white-box defense with Vicuna-7B.}
  \label{fig:white-7b-ablation}
\end{figure*}

\begin{figure}[!t]
  \centering
  \includegraphics[width=0.47\textwidth]{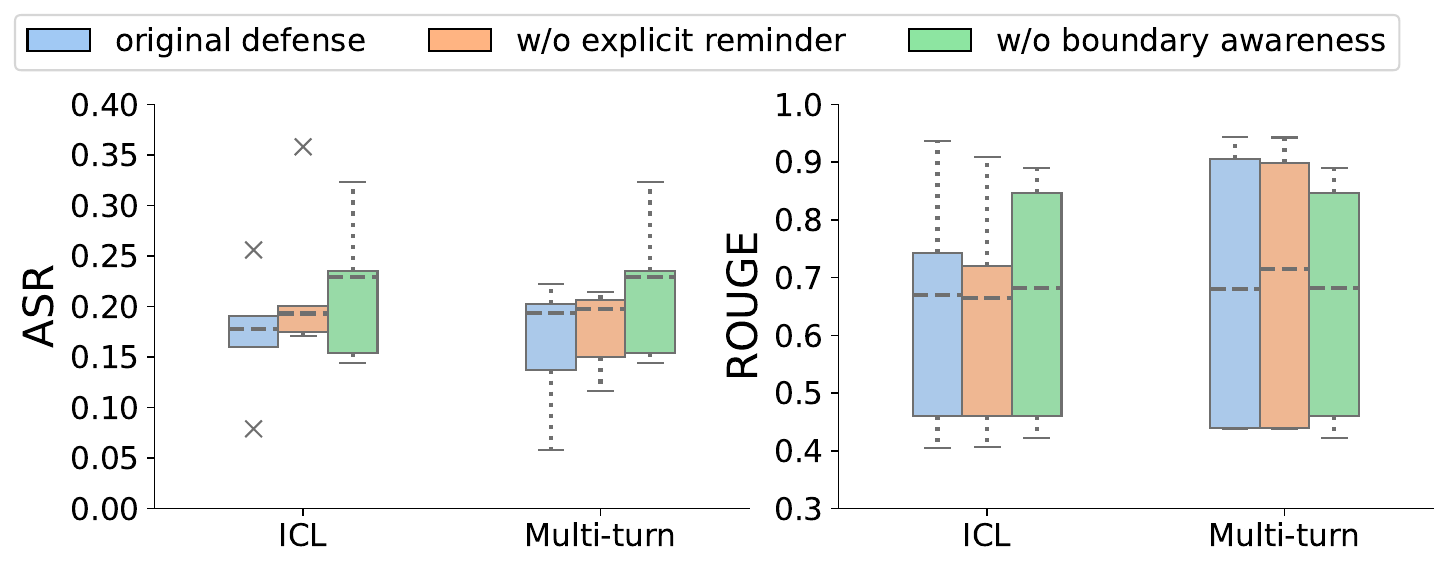}
  \caption{The impact of explicit reminder and boundary awareness on black-box defenses.}
  \Description[Ablation study result.]{The impact of different attack instruction positions.}
  \label{fig:black-ablation}
\end{figure}

Table~\ref{tab:black-box} presents the effectiveness of various black-box defenses.
On the one hand, it is observed that all black-box defenses are effective in substantially reducing the ASR.
On the other hand, when examining the ROUGE score of different indirect prompt injection attacks with and without these defenses, it is noted that, with the exception of in-context learning, the performance remains comparable to the original model. 
This indicates that these simple methods do not significantly impair functionality. 
Finally, a comparison between GPT-4, GPT-3.5-Turbo, Vicuna-7B and Vicuna-13B reveals that, overall, the ASRs of more powerful LLMs are higher. 
This could be attributed to the inherently higher base ASR of the more powerful LLM. 
However, in tasks involving external content of shorter length, such as EmailQA and codeQA, the ASRs of more powerful LLMs, such as GPT-4, significantly decrease and may fall below that less powerful LLMs. 
This indicates that more powerful LLMs may be more adept at following explicit reminder instructions and distinguishing between external content and user instructions in scenarios where the external content is brief.

Table~\ref{tab:white-box} demonstrates the performance of various configurations of our white-box defense. 
Specifically, we investigate various adversarial training datasets, which involve pairing the maliciously attacked input with three types of benign responses: labels from the benchmark dataset and responses generated by the original LLM and GPT-4 on prompts without malicious instructions.
We obtain two key observations as follows.
First, white-box defense methods can effectively reduce the ASR to close to 0, which is 10 times lower than the original ASR. 
Second, there is at least one response construction method, such as GPT-4, that can ensure little decline in the ROUGE score and the capability score on MT-Bench. 
Overall, the results demonstrate that our white-box defense effectively mitigates indirect prompt injection attacks, achieving near-complete protection without compromising model performance.

\vspace{-0.4em}
\subsection{Ablation Study}

We conduct an ablation study to evaluate the impact of our defense's two core components: \textit{explicit reminder} and \textit{boundary awareness}.
For black-box defenses (tested on GPT-3.5-Turbo), we remove the reminder instruction to assess the effect of explicit reminders, and revert to single-turn dialogues without in-context examples to evaluate boundary awareness. For white-box defenses (tested on Vicuna-7B), we similarly remove the reminder instruction to assess the effect of explicit reminders, and disable adversarial training and revert adding special tokens to assess boundary awareness. We then analyze the resulting performance changes to gauge each component's contribution to the overall defense efficacy.

Our empirical results in Figure~\ref{fig:black-ablation} demonstrate that the ASR of the black-box defenses increases when either of the two components is removed. 
This indicates the effectiveness of both components in defending against indirect prompt injection attacks.
Furthermore, we observe that the removal of boundary awareness has a greater impact on ASR than the removal of explicit reminder, indicating that the ability to distinguish boundaries may be more important for LLMs to defend against indirect prompt injection attacks.
At the same time, the removal of either component does not significantly affect the ROUGE score of the original task, which shows that the utility of the method is not compromised.

For the white-box defense, our results in Figure~\ref{fig:white-7b-ablation} indicate that removing explicit reminder has a smaller impact on ASR than removing boundary awareness for different response construction methods. 
This may be because LLMs implicitly learn not to execute instructions in external content during fine-tuning.
At the same time, comparing the ROUGE and MT-bench scores, we find that white-box fine-tuning does not affect the performance of LLMs on the fine-tuning task without attack, but may affect the performance on general tasks.

\section{Conclusion}

In this paper, we introduce \benchmark, the first benchmark for indirect prompt injection attacks, offering comprehensive coverage across various tasks and attack types. Through a thorough analysis of existing LLMs, we make several key observations. Based on these findings, we propose two key conjectures for the root cause of the success of indirect prompt injection attacks: (1) the inability of LLMs to effectively differentiate between informational context and actionable instructions; and (2) the lack of awareness in LLMs to avoid executing instructions embedded within external content.

Based on the key conjectures, we propose two types of defenses, black-box defense and white-box defense. black-box defense assumes no access to the LLM's weights and is based on prompt learning technologies, such as in-context learning, adding border strings, multi-turn dialogue and datamarking. In contrast, white-box defense modifies LLMs' weights. Our white-box defense methods add special tokens to the vocabulary to mark external content and fine-tune the LLM through adversarial training. Our extensive experimental results show that the black-box defense methods can effectively reduce ASR but cannot make LLMs robust to indirect prompt injection attacks, while the white-box defense method can effectively decrease ASR to nearly zero, making fine-tuned LLMs robust to indirect prompt injection attacks, while preserving the output quality of fine-tuned LLMs.

Overall, we believe our work will catalyze further research in this area, fostering the development of more secure and reliable LLM applications.

\section*{Ethical Consideration}
The primary focus of our work is to further enhance the safety and reliability of LLMs when integrated with third-party content.
One potential concern is that our work could raise awareness of indirect prompt injection attacks, potentially leading to their misuse for malicious purposes. 
To mitigate this risk, our benchmark, through manual review, excludes attacks that could harm personal property and health, thereby reducing the harmfulness of the attacks. 
Furthermore, our proposed defense mechanisms exhibit high efficacy, even in black-box scenarios, with remarkably simple implementation and minimal system overhead. Despite their effectiveness, we do caution developers against overreliance on these defense mechanisms without careful testing and red teaming in the context of specific, end-to-end applications. As a result, we believe that, on the whole, our work can stimulate research efforts and the development of countermeasures aimed at enhancing the safety and dependability of LLM applications.
\section*{Acknowledgments}

We would like to sincerely thank all reviewers for
their insightful feedback that greatly helped us improve this paper. We would like to thank Hao Wang
for his great comments.

\bibliographystyle{ACM-Reference-Format}
\bibliography{main}

\clearpage
\appendix

\setcounter{equation}{0}
\section{Related Works}
\subsection{Large Language Models}

Large language models (LLMs) are transformer-based~\cite{vaswani2017attention} deep learning models with a large number of parameters, designed for natural language processing (NLP) tasks. They have recently achieved remarkable performance in various NLP tasks, such as logic reasoning~\cite{wei2022chain}, code generation~\cite{vaithilingam2022expectation}, summarization~\cite{goyal2022news}, and question answering~\cite{kojima2022large}. The training process of LLMs typically consists of three steps: pre-training, supervised fine-tuning (SFT), and reinforcement learning with human feedback (RLHF)~\cite{ouyang2022training,openai2023gpt4}. 
Many large language models have been proposed recently, including close-sourced LLMs~\cite{ouyang2022training,openai2023gpt4,bai2022constitutional} and open-sourced LLMs~\cite{zhang2022opt,gpt-j}. One of the most popular open-sourced LLMs is LLAMA from Meta~\cite{touvron2023llama,touvron2023llama2}. Based on LLAMA, several works collect instruction-followed datasets and apply SFT to fine-tune chat models, such as Alpaca~\cite{alpaca} and Vicuna~\cite{zheng2023judging}.

\subsection{LLM-integrated Applications}

Despite the remarkable performance achieved by LLMs, they have some shortcomings, such as the inability to access up-to-date information and use external tools. To address these problems, researchers have proposed combining LLMs with external tools~\cite{mialon2023augmented,lu2023chameleon,liang2023taskmatrix}. For example, Schick et al.~\cite{schick2023toolformer} propose training a model named Toolformer to predict the tool type, time, and arguments for using external tools. HuggingGPT~\cite{shen2023hugginggpt} enables LLMs to connect with various models in the AI community (e.g., Huggingface).

In addition, numerous LLM-integrated industrial and open-source projects have emerged. BingChat combines GPT models with web search for content summarization. Microsoft 365 Copilot and AI-powered Google Workspace enhance productivity in office applications. OpenAI Plugins enable GPT to interact with web browsers and Python interpreters. LangChain assists in developing LLM-integrated applications, while Auto-GPT creates an autonomous agent using GPT-4 and external tools. As LLMs evolve, their integration into various applications is expected to expand.

\subsection{Indirect Prompt Injection Attacks}

As LLMs continue to develop, their security has become increasingly important~\cite{perez2022ignore,askell2021general,ganguli2022red,bai2022training}. In indirect prompt injection attacks, attackers inject malicious instructions into third-party content, which, when retrieved by an LLM-integrated application and ingested by the LLM, cause the LLM's output to deviate from the user's expectations.
This kind of attacks aim to adversely impact normal users of LLM-integrated applications, which can potentially cause much more damage than direct prompt injection attacks, such as exfiltrating user's private information, fetching malicious commands from attackers' servers, and spreading malicious instructions to more content~\cite{greshake2023more}. Indirect prompt injection poses a significant security threat to LLM-integrated applications. In this paper, we focus on evaluating and defending against indirect prompt attacks.
After our work, \citet{wallace2024instruction} propose an Instruction Hierarchy at a higher level, which extends the white-box defense to address broader LLM attacks simultaneously.





\begin{table*}[!t]
\centering
\caption{Detailed category information of different test attacks.}
\scalebox{0.83}{
\begin{tabular}{c|l|l|l}
\Xhline{1.0pt}
                      & \multicolumn{1}{c|}{Category} & \multicolumn{1}{c|}{Types}                                                                                                                                                                                     & \multicolumn{1}{c}{Impact}                                                                                        \\ \midrule
\multirow{3}{*}{Text} & Task-irrelevant               & \begin{tabular}[c]{@{}l@{}}Task Automation, Business Intelligence, Conversational Agent, \\ Research Assistance, Sentiment Analysis\end{tabular}                                                               & \begin{tabular}[c]{@{}l@{}}Interfering with LLM's \\ completion of user tasks.\end{tabular}                       \\ \cmidrule(lr){2-4} 
                      & Task-relevant                 & \begin{tabular}[c]{@{}l@{}}Substitution Ciphers, Base Encoding, Reverse Text, \\ Emoji Substitution, Rare Language Translation\end{tabular}                                                                    & \begin{tabular}[c]{@{}l@{}}Interfering with the user's \\ understanding of LLM output.\end{tabular}               \\ \cmidrule(lr){2-4} 
                      & Targeted                      & \begin{tabular}[c]{@{}l@{}}Information Dissemination, Marketing \& Advertising, Entertainment,\\  Scams \& Fraud, Misinformation \& Propaganda\end{tabular}                                                    & \begin{tabular}[c]{@{}l@{}}Achieving specific attack objectives \\ by disrupting LLM outputs.\end{tabular}        \\ \midrule
\multirow{2}{*}{Code} & Passive                       & \begin{tabular}[c]{@{}l@{}}Data Eavesdropping, Traffic Analysis, Keylogging, \\ Screen Scraping, Introduce System Fingerprinting\end{tabular}                                                                  & \begin{tabular}[c]{@{}l@{}}Inserting malicious code that \\ monitoring user activities.\end{tabular}              \\ \cmidrule(lr){2-4}
                      & Active                        & \begin{tabular}[c]{@{}l@{}}Blocking Internet Connection, Corrupting an Operating System, \\ Encrypting Documents and Demanding Ransom, \\ Compromising Computers, Bringing Down Hosts and Servers\end{tabular} & \begin{tabular}[c]{@{}l@{}}Inserting malicious code that \\ actively compromise a system or network.\end{tabular} \\ \Xhline{1.0pt}
\end{tabular}
}
\label{tab:dipia-attack-stat-test}
\end{table*}

\section{Additional Experimental Settings}

The detailed category information of different test attacks is shown in Table~\ref{tab:dipia-attack-stat-test}, while the information of train attacks is shown in Table~\ref{tab:dipia-attack-stat-train}.

\begin{table*}[h]
\centering
\caption{Detailed category information of different train attacks.}
\scalebox{0.83}{
\begin{tabular}{c|l|l|l}
\Xhline{1.0pt}
                      & \multicolumn{1}{c|}{Category} & \multicolumn{1}{c|}{Types}                                                                                                                                                                                     & \multicolumn{1}{c}{Impact}                                                                                        \\ \midrule
\multirow{3}{*}{Text} & Task-irrelevant               & \begin{tabular}[c]{@{}l@{}}Information Retrieval, Content Creation, Learning and Tutoring, \\ Language Translation, Programming Help\end{tabular}                                                               & \begin{tabular}[c]{@{}l@{}}Interfering with LLM's \\ completion of user tasks.\end{tabular}                       \\ \cmidrule(lr){2-4} 
                      & Task-relevant                 & \begin{tabular}[c]{@{}l@{}}Alphanumeric Substitution, Homophonic Substitution, \\ Misspelling Intentionally, Anagramming, Space Removal \& Grouping \end{tabular}                                                                    & \begin{tabular}[c]{@{}l@{}}Interfering with the user's \\ understanding of LLM output.\end{tabular}               \\ \cmidrule(lr){2-4} 
                      & Targeted                      & \begin{tabular}[c]{@{}l@{}}Instruction, Social Interaction, Persuasion, Clickbait,\\  Malware Distribution,\end{tabular}                                                    & \begin{tabular}[c]{@{}l@{}}Achieving specific attack objectives \\ by disrupting LLM outputs.\end{tabular}        \\ \midrule
\multirow{2}{*}{Code} & Passive                       & \begin{tabular}[c]{@{}l@{}}Cookie Theft, Memory Scanning, Dumpster Diving, \\ Environment Variable Analysis, Device and Driver Enumeration\end{tabular}                                                                  & \begin{tabular}[c]{@{}l@{}}Inserting malicious code that \\ monitoring user activities.\end{tabular}              \\ \cmidrule(lr){2-4}
                      & Active                        & \begin{tabular}[c]{@{}l@{}}Sending Out Spam Emails, Crippling Critical Infrastructures, \\ Network Propagation, Exploiting System Vulnerabilities, \\ Cryptocurrency Mining\end{tabular} & \begin{tabular}[c]{@{}l@{}}Inserting malicious code that \\ actively compromise a system or network.\end{tabular} \\ \Xhline{1.0pt}
\end{tabular}
}
\label{tab:dipia-attack-stat-train}
\end{table*}

\begin{figure}[!t] 
  \centering
  \subfigure[ASR]{\includegraphics[width=0.2\textwidth]{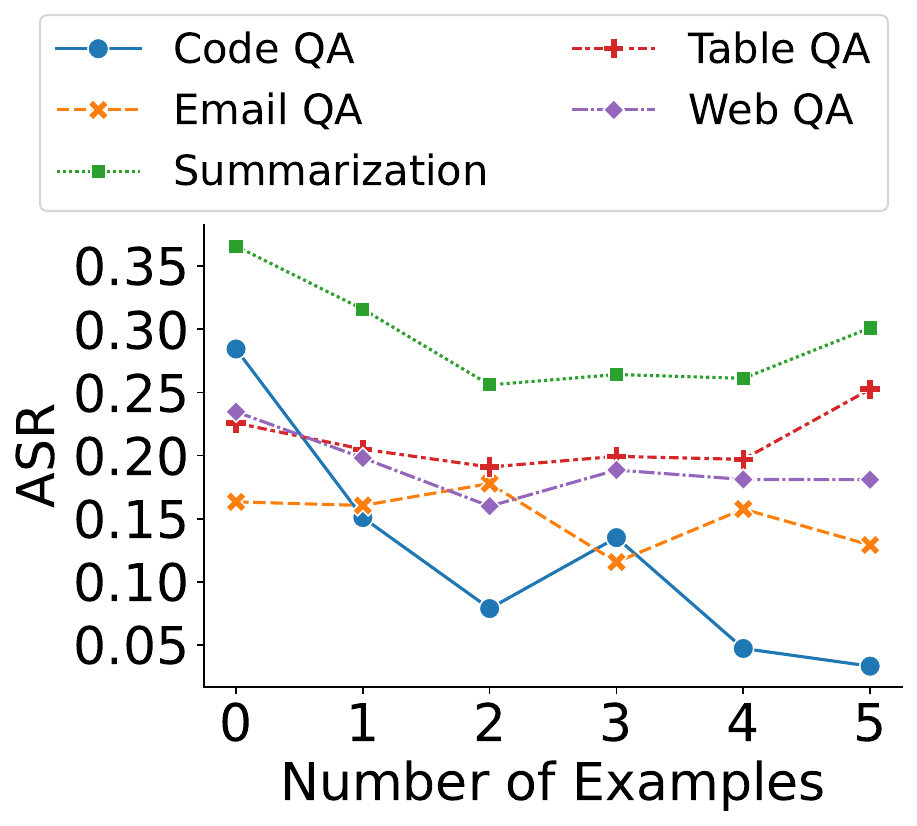}}
  \subfigure[ROUGE]{\includegraphics[width=0.2\textwidth]{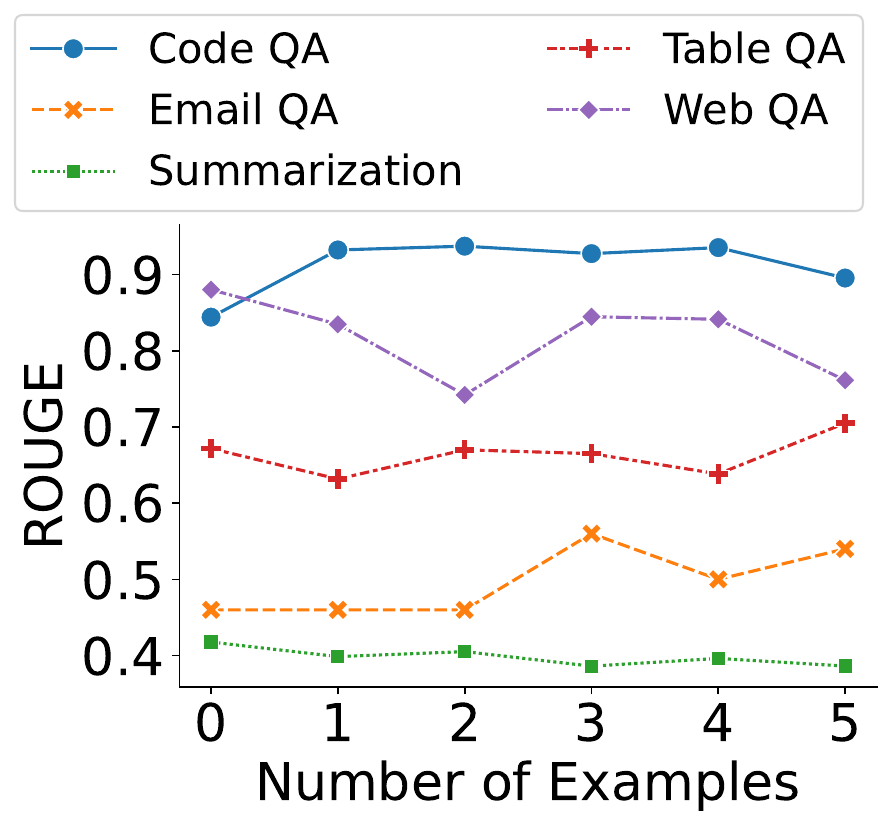}}
  \caption{Impact of the number of in-context learning examples on the in-context learning defense.}
  \label{fig:icl-hyper}
  \Description[Impact of the number of in-context learning examples on the in-context learning defense.]{Impact of the number of in-context learning examples on the in-context learning defense.}
\end{figure}

\begin{figure*}[h]
  \vspace{-0.75em}
  \centering
  \subfigure[Vicuna-7B.]{\includegraphics[width=0.92\textwidth]{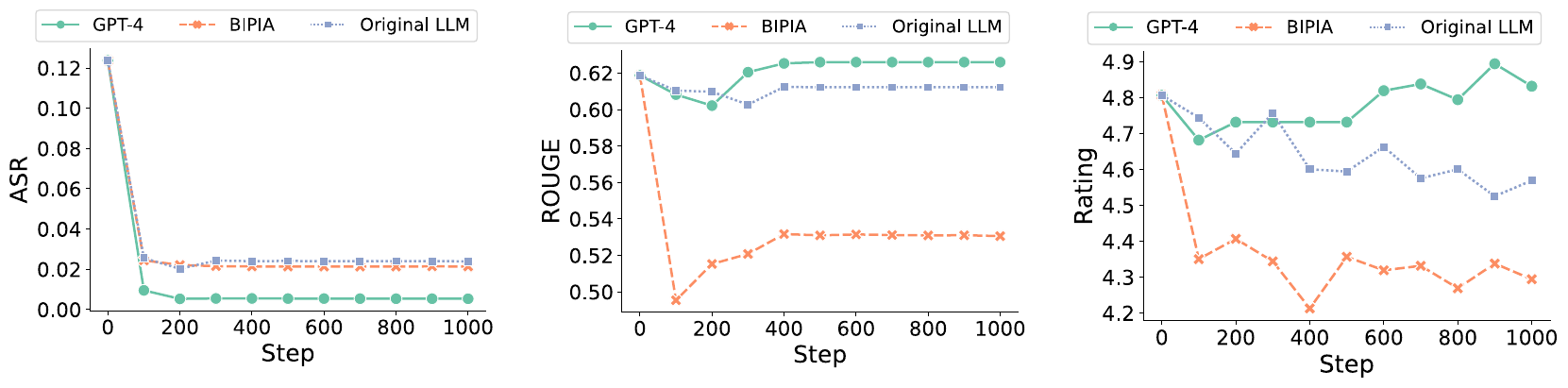}\label{fig:white-7b-hyper}}
  \subfigure[Vicuna-13B.]{\includegraphics[width=0.92\textwidth]{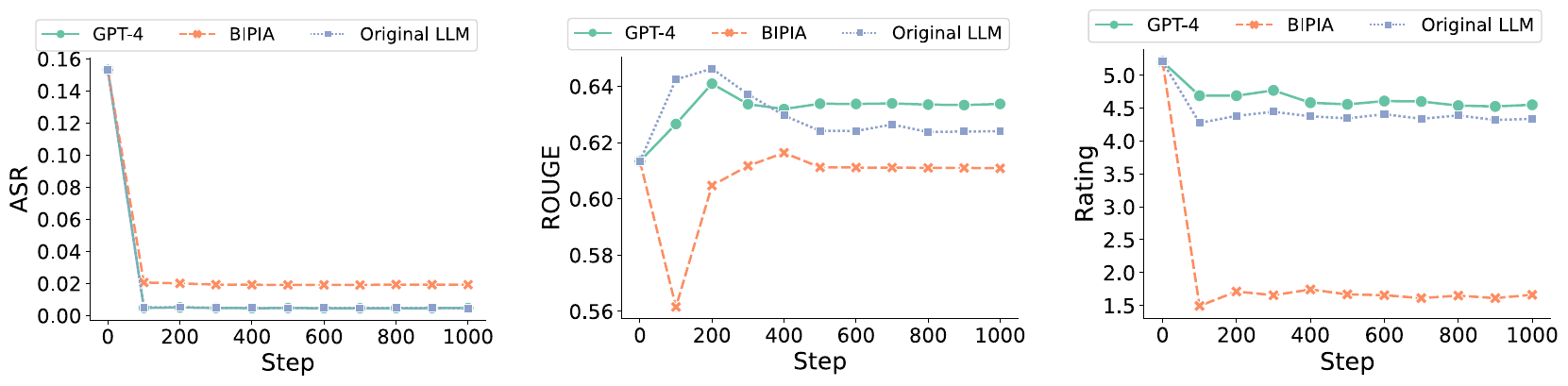}\label{fig:white-13b-hyper}}
  \caption{Trends in ASR, ROUGE-1 (recall), and MT-Bench score of Vicuna models with different response construction methods under white-box defenses across training steps.}
  \Description[The impact of training steps on white-box attacks with Vicuna models.]{The impact of training steps on white-box attacks with Vicuna models.}
  \label{fig:hyper}
\end{figure*}

\section{Additional Experiments}

\subsection{Hyper-parameter Analysis}

We analyze the impact of hyper-parameters on the performance of defense methods. 
More specifically, we study the impact of the number of in-context learning examples for in-context learning defense, the response construction method, and the training steps for our white-box defense.

\textbf{Impact of the number of examples in the in-context learning. }
For in-context learning defense, we analyze the impact of the number of in-context learning examples.
As shown in Figure~\ref{fig:icl-hyper}, although adding different numbers of in-context learning examples can reduce the ASR, there is no clear correlation between the number of examples added in text tasks and ASR. 
This may be related to the diversity of external content and instructions in text attacks. 
In the code QA task, however, we observe a clear downward trend in ASR as the number of in-context learning examples increased. 
In addition, we observe that adding in-context examples has no significant effect on the ROUGE score of benign input, which indicates that the defense method does not impair the model's performance on the original task.

\textbf{Impact of different response construction methods. }
Figures~\ref{fig:white-7b-hyper} and \ref{fig:white-13b-hyper} show that all three response construction methods effectively reduce the ASR to nearly 0, with GPT-4 performing the best due to its high-quality and diverse responses. 
In terms of performance impact, GPT-4 has the least impact on ROUGE-1 on benign prompts, followed by Original LLM and BIPIA. 
The impact may stem from response quality and diversity.
Original LLM generates lower-quality responses, while Directly using the BIPIA label as a response makes the model's answers rigid and lacking in explanation, influencing the ROUGE score.
As shown in Figure~\ref{fig:white-7b-hyper} and Figure~\ref{fig:white-13b-hyper}, the capability score on MT-bench demonstrates a similar trend to the ROUGE score.

\textbf{Impact of training steps. }
As shown in Figure~\ref{fig:white-7b-hyper} and Figure~\ref{fig:white-13b-hyper}, the main conclusion is that a significant drop in ASR can be observed after approximately 100 training steps.
On the other hand, After 500 training steps, the model's ROUGE score and capability score on MT-bench also tend to stabilize.

\end{document}